\documentclass{article}
\usepackage{graphicx,geometry,amssymb,amsmath, subcaption} 

\title{Extending confidence calibration to generalised measures of variation}
\author{Andrew Thompson and Vivek Desai\\
{\small National Physical Laboratory, Hampton Road, Teddington, TW11 0LW, United Kingdom}\\
    {\small \texttt{vivek.desai@npl.co.uk}}}

\date{February 2026}

\begin{document}

\newtheorem{thm}{Theorem}
\newcommand{\ra}{\rightarrow}

\maketitle

\begin{abstract}
We propose the Variation Calibration Error (VCE) metric for assessing the calibration of machine learning classifiers. The metric can be viewed as an extension of the well-known Expected Calibration Error (ECE) which assesses the calibration of the maximum probability or \emph{confidence}. Other ways of measuring the variation of a probability distribution exist which have the advantage of taking into account the full probability distribution, for example the Shannon entropy. We show how the ECE approach can be extended from assessing confidence calibration to assessing the calibration of any metric of variation. We present numerical examples upon synthetic predictions which are perfectly calibrated by design, demonstrating that, in this scenario, the VCE has the desired property of approaching zero as the number of data samples increases, in contrast to another entropy-based calibration metric (the UCE) which has been proposed in the literature.
    
\end{abstract}

\section{Introduction}\label{sec:intro}

The importance of assessing the trustworthiness of machine learning models through uncertainty quantification is now widely recognised -- for example the EU AI Act indirectly mandates uncertainty quantification by requiring transparency and risk management~\cite{deuschel2024role} -- and a natural way to do this is to assign probabilities to each of the classes for a given prediction. It is also important that these probabilities give a reliable indication of the uncertainty in the model's predictions. Many classification models do output scores for the classes, but in the case of deep neural networks these scores are often poorly calibrated and therefore not meaningful as statements of probability~\cite{guo2017calibration}.

Calibration analysis is a technique for evaluating the extent to which the class probabilities returned by the model are reliable. In simple terms, a well-calibrated model is one for which the class probabilities produced by the model match empirical observations. Various metrics which capture calibration of classifiers have now been proposed; see for example~\cite{silva2023classifier} for a survey. 

Attention is often given in classification models to the \emph{predicted class}, defined to be the class with largest returned probability. The corresponding maximum probability value is often referred to as the \emph{confidence} of the prediction. Many calibration metrics adopt the approach of \textit{confidence calibration}, with a well-calibrated model being viewed as one for which the prediction confidences match observed proportions of correct predictions. Many confidence calibration metrics, such as Expected Calibration Error (ECE)~\cite{naeini2015obtaining}, involve binning the data according to outputted confidence values, calculating the observed proportions of correct classifications within each bin, and then combining the calibration errors for each bin into a single metric.

Confidence is one example of a summary statistic which quantifies the \emph{variation} of a probability distribution, but other variation metrics also exist (see Section~\ref{rationale} for further details). One of the drawbacks of confidence as a variation metric (in the case of more than two classes) is that it only takes into account the probability corresponding to the most likely class, and not the values of the other probabilities.

In this paper, we introduce Variation Calibration Error (VCE) as a way of extending the concept of confidence calibration to calibration of general variation metrics for probability distributions. We achieve this extension by comparing the \emph{order statistics} of a probability prediction -- that is to say, the probabilities ordered by size -- with the observed proportions of the \emph{rankings} assigned to the true class. We describe the approach in detail in Section~\ref{sec:VCE}, showing how it can be viewed as a natural extension to confidence calibration.

One possible choice of variation metric which takes into account the entire probability distribution is entropy. A calibration metric called Uncertainty Calibration Error (UCE) was proposed in~\cite{laves2019well} in which the \emph{normalised Shannon entropy} (we will refer to it henceforth as simply entropy) of the returned probability distribution is compared against the observed proportion of incorrect predictions (or \emph{misclassification rate}). A binning approach is also taken for UCE, with data binned according to the entropy of outputted probabilities and then compared against observed misclassification rates per bin.

It is not intuitively obvious why the predicted entropies and the misclassification rate might be expected to be correlated for a well-calibrated model. Indeed, we argue in Section~\ref{sec:VCE} that our calibration approach with entropy as the chosen variation metric offers a more intuitive approach to calibrating entropy. In Section~\ref{sec:experiments}, we present some simple numerical experiments upon synthesised predictions whose probability distribution is perfectly calibrated by design. We show that both the ECE and the VCE approach zero as the number of samples increases, whereas the UCE does not have this desirable property. These experiments provide evidence that the more intuitive VCE is a more reliable indicator of calibration quality than the UCE.

We emphasise that our proposed approach is a generalisation of confidence calibration. By choosing confidence as the variation metric, we recover confidence calibration as a special case. Similarly, we show in Section~\ref{sec:VCE} that ECE is a special case of VCE.

Before proceeding, we note that other approaches that extend beyond confidence calibration in the multiclass classification setting have been proposed. One alternative approach is \emph{classwise calibration}~\cite{zadrozny2002transforming}, in which the problem is reframed as a series of one-versus-rest binary classification tasks, the calibration of each of which is then assessed separately. An entirely different approach to binning-based metrics for multiclass calibration was proposed in~\cite{widmann2019calibration} based on matrix-valued kernels.

\section{Variation Calibration Error (VCE)}\label{sec:VCE}

\subsection{The rationale}\label{rationale}

In confidence calibration, the largest returned probability is compared with the observed proportion of cases in which the true class is the one with largest predicted probability. We may also compare the second-largest returned probability with the observed proportion of cases in which the true class is the one with second-largest predicted probability. We may proceed analogously for each of the ranked returned probabilities, comparing with observed proportions of cases in which the true class is the one with the corresponding ranking. In terms of comparing the entire probability distribution, then, we compare the reverse-ordered predicted probability distribution with the \emph{empirical probability distribution consisting of the observed proportions of the rankings of the true class}. 

We may now compare the two probability distributions using any variation metric for probability distributions of our choosing. Various metrics capturing the variation of a discrete probability distribution have been proposed, including Wilcox's Variation Ratio (WVR), entropy and the Index of Qualitative Variation (IQR). In this paper, we will focus on the use of entropy, which is commonly used in the ML community for summarising classification uncertainty. We refer the interested reader to~\cite{wilcox1973indices,bilson2025metrological} for further details on WVR, IQR and other variation metrics for probability distributions.

Given $C$ classes, we follow~\cite{bilson2025metrological} in defining the entropy $\mathcal{H}(p)$ of a probability vector 
$$p=\begin{bmatrix}p_1&\cdots&p_C\end{bmatrix}^{\top}$$
to be
$$\mathcal{H}(p):=-\sum_{c=1}^C p_c \log_C p_c.$$

\subsection{Precise definition}

We next show more precisely how one popular confidence calibration metric, the ECE~\cite{naeini2015obtaining}, can be extended to a metric which assesses more generalised variation calibration, which we refer to as VCE. We note at the outset that ECE is only one of the metrics proposed for confidence calibration, and the broad approach described in Section~\ref{sec:intro} could also be used to extend other confidence calibration metrics, such as Adaptive Calibration Error (ACE)~\cite{nixon2019measuring}.

As mentioned in Section~\ref{sec:intro}, ECE involves binning the data according to outputted confidence values, in this case using equal-width bins. Given $N$ samples indexed by $i=1,\ldots,N$, we write $\hat{p}_i$ for the largest probability value (confidence) for sample $i$. Given number of bins $M$, we partition the confidence values into $M$ equally-spaced bins. Then, provided $|B_m|\neq 0$, we define
$$\mathrm{conf}(B_m):=\frac{1}{|B_m|}\sum_{i\in B_m} \hat{p}_i,$$
and we define
$$\textrm{acc}(B_m):=\frac{1}{|B_m|}\sum_{i\in B_m}\mathbf{1}(\hat{y}_i=y_i),$$
where $y_i$ and $\hat{y}_i$ are the true and predicted class labels respectively for
sample $i$. The ECE is then given by
$$\mathrm{ECE}:=\sum_{m=1}^M\frac{|B_m|}{N}|\textrm{acc}(B_m)-\textrm{conf}(B_m)|.$$

Like other calibration metrics that are based on confidence, ECE suffers from the drawback of not taking into account the full probability distribution. To extend the ECE to generalised measures of variation, we choose a variation metric $\mathcal{V}$ and generalise both $\mathrm{conf}(B_m)$ and $\mathrm{acc}(B_m)$. Given number of bins $M$, redefine the set $B_m$ to be
$$B_m:=\left\{i:\mathcal{V}(p_i)\in\left(\frac{m-1}{M},\frac{m}{M}\right]\right\}.$$
Given $C$ classes, write $q_i$ for the length-$C$ vector consisting of the reverse-ordered probabilities for sample $i$ and, provided $|B_m|\neq 0$, define
$$\mathcal{P}(B_m):=\mathcal{V}\left(\frac{1}{|B_m|}\sum_{i\in B_m} q_i\right).$$
In the case of equal probabilities in $q_i$, the above definition is not well-defined. However, the definition can be extended in this case by using any tie-breaking scheme to arbitrarily enforce an ordering. For sample $i$, write $r_i$ for the length-$C$ vector whose $c^{\mathrm{th}}$ entry is
$$(r_i)_c:=\mathbf{1}(y_i=k_{ic}),$$
where $k_{ic}$ is the class corresponding to the $c^{\mathrm{th}}$ entry of $q_i$, and then define, provided $|B_m|\neq 0$, 
$$\mathcal{O}(B_m):=\mathcal{V}\left(\frac{1}{|B_m|}\sum_{i\in B_m} r_i\right).$$
Here $\mathcal{P}$ and $\mathcal{O}$ stand for \emph{predicted variation} and \emph{observed variation} respectively. Analogous to the ECE, the VCE is then given by
$$\mathrm{VCE}:=\sum_{m=1}^M\frac{|B_m|}{N}|\mathcal{O}(B_m)-\mathcal{P}(B_m)|.$$
The definitions of $\mathcal{P}(B_m)$ and $\mathcal{O}(B_m)$ given above are not well-defined if the set $B_m$ is empty. However, in this case, we may set both $\mathcal{P}(B_m)$ and $\mathcal{O}(B_m)$ to arbitrary finite values, since they will make no contribution to the VCE due to subsequent multiplication by $|B_m|=0$.

It is straightforward to check that the VCE reduces to the ECE when the variation metric is chosen to be confidence, that is, the value of the largest probability.

\section{Numerical experiments}\label{sec:experiments}

\begin{figure*}[t!]
    \centering
    \includegraphics[width=0.45\linewidth]{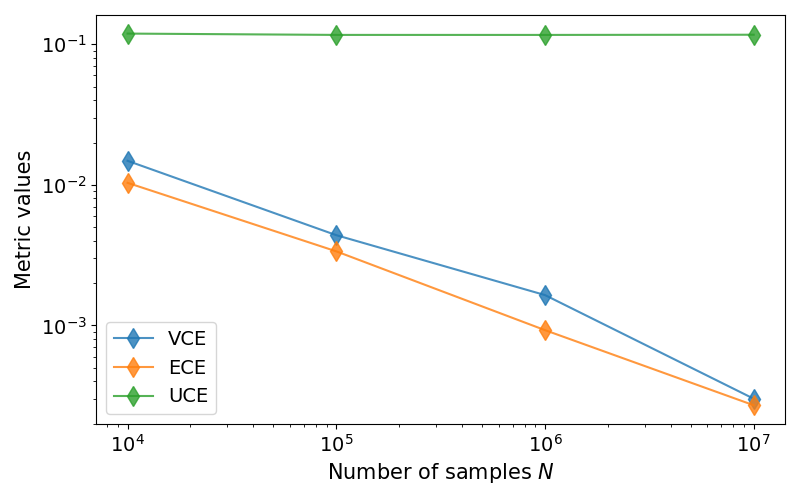}
    \includegraphics[width=0.45\linewidth]{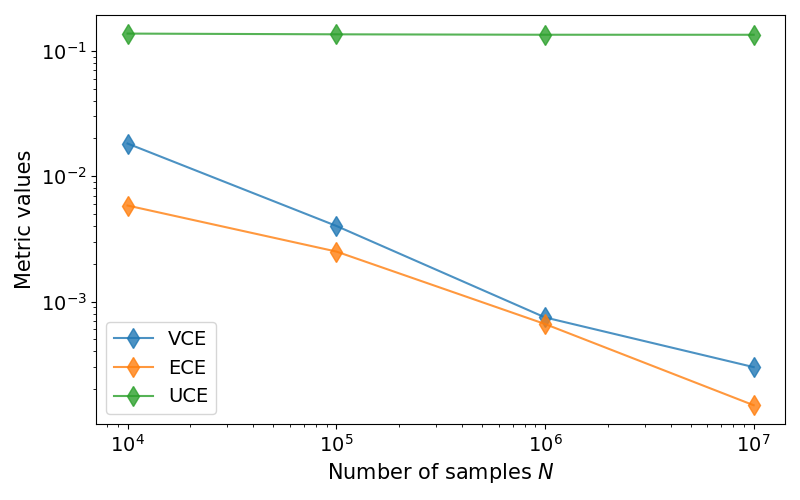}
    \includegraphics[width=0.45\linewidth]{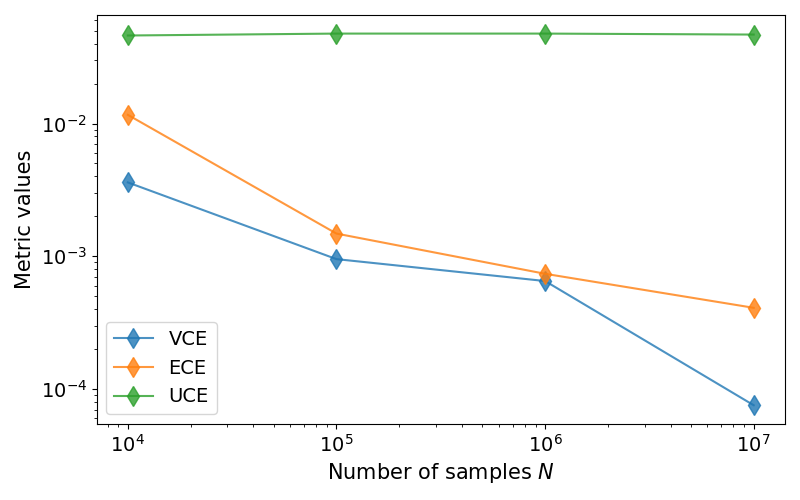}
    \includegraphics[width=0.45\linewidth]{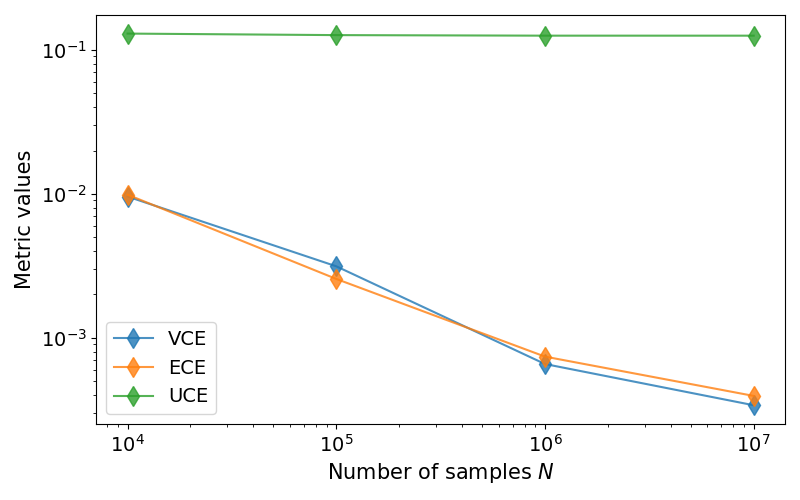}
    \caption{Metric results for the equal-width binning strategy for the VCE, ECE, and UCE metrics. Results are shown for the $3$-class (top) and $10$-class (bottom) classification problems, with two different sets of $\alpha$ parameters (equally weighted (left) and heavily skewed (right)), across four different numbers of samples.}
    \label{fig:EW_results}
\end{figure*}

\begin{figure*}[ht]
    \centering
    \includegraphics[width=0.32\linewidth]{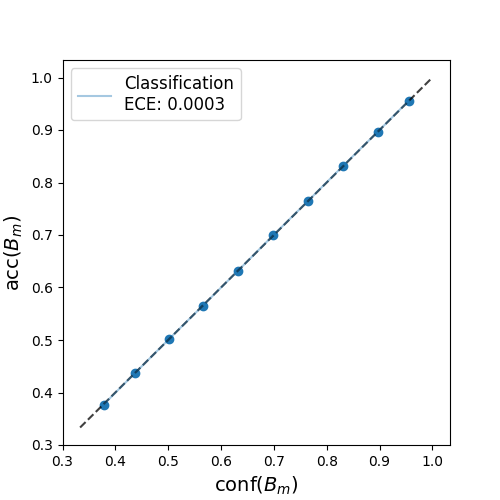}
    \includegraphics[width=0.32\linewidth]{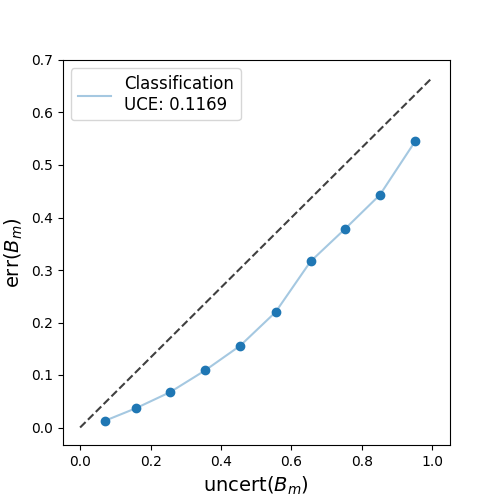}
    \includegraphics[width=0.32\linewidth]{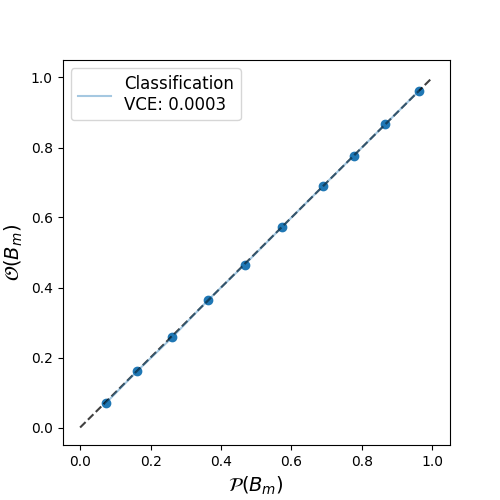}
    \caption{Reliability diagrams for the ECE (left), UCE (middle) and VCE (right) for an example experiment, with $C=3$, equally-weighted $\alpha$ parameters and $N=10^{7}$. We adopt an equal-width binning strategy for these results. The black dashed line indicates perfect calibration. Respective metric values are shown in the legend of each plot.}
    \label{fig:EW_reliability_diagrams}
\end{figure*}

We generate predictions for a $C$-class classification task which are perfectly calibrated by design. For each sample $i\in\{1,\ldots,N\}$, we sample a predicted probability vector 
$$p_i=\begin{bmatrix}(p_i)_1&\cdots&(p_i)_C\end{bmatrix}^{\top}$$
from the same Dirichlet distribution with parameters $\alpha=(\alpha_1,\ldots,\alpha_C)$. The $\alpha$ parameters determine the relative frequency of the classes. For each sample $i\in\{1,\ldots,N\}$, we then use these predicted probabilities to draw a ``true'' class; in other words, $y_i=c$ with probability $(p_i)_c$.

For each experiment, characterised by a choice of $\alpha$, we calculate ECE~\cite{naeini2015obtaining}, UCE~\cite{laves2019well}, and VCE with entropy as the variation metric, for varying choices of $N$.

We already introduced the ECE and the VCE in Section~\ref{sec:VCE}. When applying the ECE to a $C$-class classification task, the confidence values are restricted to the interval $[1/C,1]$, and so we define $M$ bins which cover this interval. The UCE~\cite{laves2019well} is defined to be
$$\mathrm{UCE}:=\sum_{m=1}^M\frac{|B_m|}{N}|\textrm{err}(B_m)-\textrm{uncert}(B_m)|,$$
where $B_m$ is defined as in Section~\ref{sec:VCE}, and where
$$\textrm{err}(B_m):=1-\textrm{acc}(B_m)=\frac{1}{|B_m|}\sum_{i\in B_m}\mathbf{1}(\hat{y}_i\neq y_i)$$
and
$$\textrm{uncert}(B_m):=\frac{1}{|B_m|}\sum_{i\in B_m} \mathcal{H}(p_i).$$

We fix the number of bins to be $M=10$ for all experiments. We also investigate the impact on metric results of replacing equal-width bins with adaptive \emph{equal-frequency} bins (same number of samples in each bin). 

\subsection{Equal-width results}\label{sec:EW_results}

Figure \ref{fig:EW_results} shows the results of evaluating the different calibration metrics for different perfectly calibrated scenarios with varying numbers of classes and varying parametrisations of the Dirichlet distribution. We test two choices for the number of classes: $C \in\{3, 10\}$. For each choice of number of classes, we also test two different choices for the $\alpha$ parameters: equal weight distributed across all classes ($\alpha_c=1$ for $c=1,\ldots,C$), and a weighting heavily skewed towards one of the classes ($\alpha_1=10$ and $\alpha_c=1$ for $c=2,\ldots,C$.  Each metric is evaluated per scenario for four different numbers of samples $N \in \{10^{4}, 10^{5}, 10^{6}, 10^{7}\}$, and we use equal-width binning as described in Section \ref{sec:VCE}. 

Figure \ref{fig:EW_results} demonstrates that the VCE metric exhibits the desirable behaviour of decaying to zero as the number of samples $N$ increases for perfectly calibrated scenarios. In the case of UCE, a noise floor persists despite the use of perfectly calibrated data, even under increasing number of samples. The ECE and VCE behave similarly, decreasing monotonically with increasing $N$. These observations hold for different sets of the concentration parameter $\alpha$, even though the Dirichlet distribution is significantly peaked near the simplex vertex corresponding to the highest $\alpha$ parameter. 

Figure \ref{fig:EW_reliability_diagrams} shows the reliability diagrams for each metric for a specific experiment: three classes, equal $\{\alpha_i\}$ and $N=10^7$. For each reliability diagram, the dashed line represents perfect calibration according to the definition of each metric. We see that the ECE and VCE metrics exhibit almost perfect calibration across all bins, whereas the UCE deviates from the ideal calibration relationship. These reliability diagrams highlight how the UCE exhibits systematic issues with its definition of calibration, which the VCE avoids. 

\subsection{Equal-frequency results}\label{sec:EF_results}

\begin{figure*}[t!]
    \centering
    \includegraphics[width=0.45\linewidth]{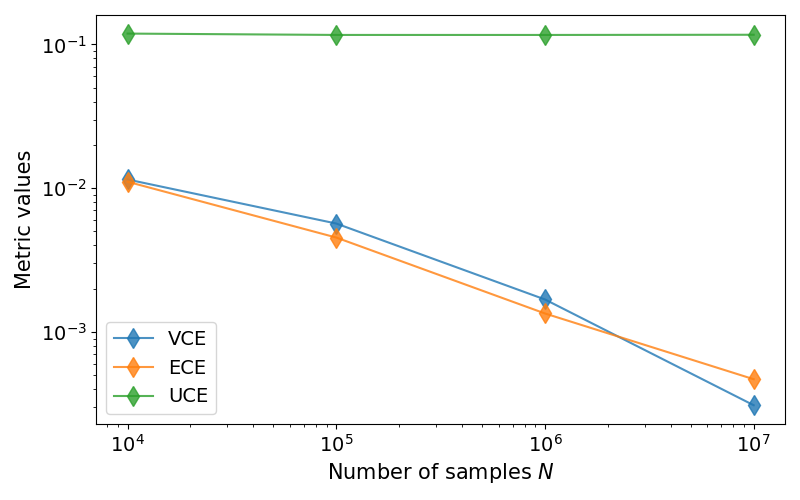}
    \includegraphics[width=0.45\linewidth]{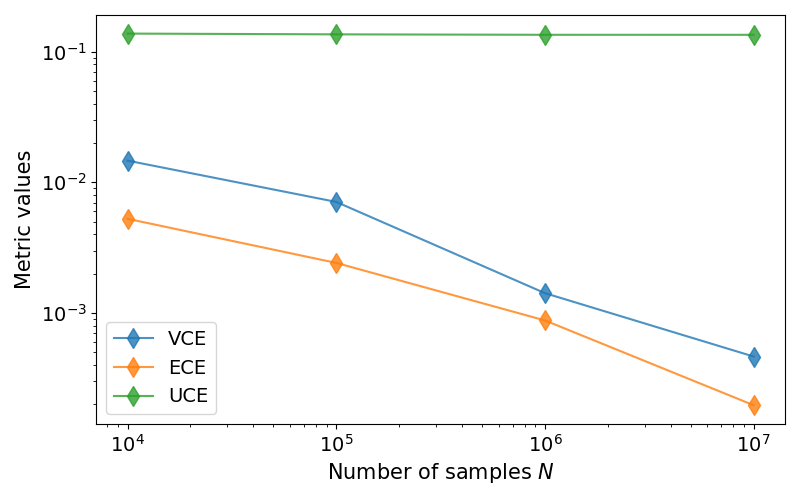}
    \includegraphics[width=0.45\linewidth]{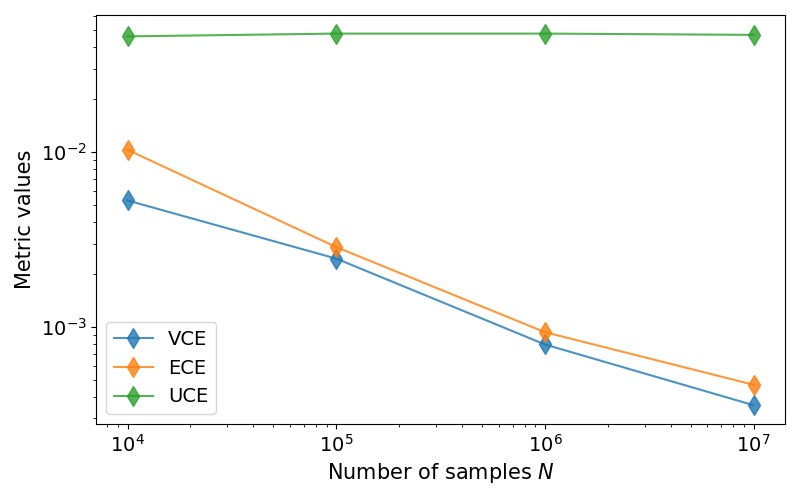}
    \includegraphics[width=0.45\linewidth]{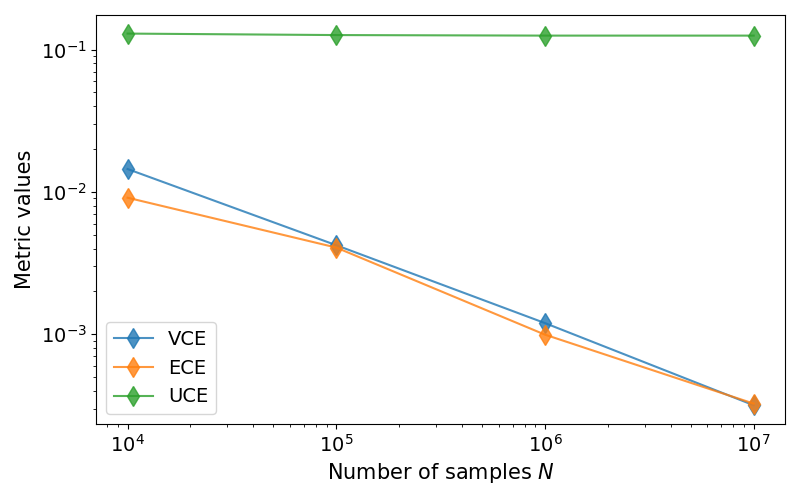}
    \caption{Metric results for the equal-frequency binning strategy for the VCE, ECE, and UCE metrics. Results are shown for the 3-class (top) and 10-class (bottom) classification problems, with two different sets of $\alpha$ parameters (equally weighted (left) and heavily skewed (right)), across four different numbers of samples.}
    \label{fig:EF_results}
\end{figure*}

Figure \ref{fig:EF_results} shows the results of evaluating the different calibration metrics for the same four perfectly calibrated scenarios, with a modified binning strategy in which all bins $B_m$ contain the same number of samples. We examine the behaviour of the metrics for each scenario with the same four choices of number of samples $N \in \{10^{4}, 10^{5}, 10^{6}, 10^{7}\}$. 

Just as in the case of equal-width bins, we observe that the ECE and VCE both have the desirable property of decaying to zero as the number of samples $N$ increases, whereas a noise floor persists for UCE even as the number of samples increases.

\section{Conclusion}\label{sec:summary}

We have proposed the VCE which assesses the calibration of generalised variation metrics for ML classification predictions. In particular this allows for an extension of the ECE to variation metrics which take into account the full probability distribution. We have provided numerical evidence that the metric with entropy behaves as desired for predictions which are perfectly calibrated by design, in contrast to the UCE. A remaining question is whether these numerical observations can be supported by theoretical results, either for the case of entropy or for a wider class of variation metrics, and this is left as future work.

\section*{Acknowledgments}

This work was carried out as part of the QUMPHY project (22HLT01), which has received funding from the European Partnership on Metrology, co-financed from the European Union’s Horizon Europe Research and Innovation Programme and by the Participating States. Funding for NPL was provided by Innovate UK under the Horizon Europe Guarantee Extension, grant number 10084125. The authors wish to thank Samuel Bilson (NPL) for reviewing the report and providing useful feedback.

\bibliographystyle{plain}
\bibliography{references}

@article{bilson2025metrological,
  title={A metrological framework for uncertainty evaluation in machine learning classification models},
  author={Bilson, Samuel and Cox, Maurice and Pustogvar, Anna and Thompson, Andrew},
  year={2025},
  volume = {62},
  number = {6},
  journal = {Metrologia} }

@inproceedings{laves2019well,
  title={Well-calibrated Model Uncertainty with Temperature
Scaling for Dropout Variational Inference},
  author={Laves, Max-Heinrich and Ihler, Sontje and Kortmann, Karl-Philipp and Ortmaier, Tobias}, 
  year = {2019},
booktitle = {4th workshop on Bayesian Deep Learning (NeurIPS 2019)},
address = {Vancouver, Canada}
}

@inproceedings{nixon2019measuring,
  title={Measuring Calibration in Deep Learning.},
  author={Nixon, Jeremy and Dusenberry, Michael W and Zhang, Linchuan and Jerfel, Ghassen and Tran, Dustin},
  booktitle={CVPR workshops},
  volume={2},
  year={2019}
}

@inproceedings{naeini2015obtaining,
  title={Obtaining well calibrated probabilities using {B}ayesian binning},
  author={Naeini, Mahdi Pakdaman and Cooper, Gregory and Hauskrecht, Milos},
  booktitle={Proceedings of the AAAI Conference on Artificial Intelligence},
  volume={29},
  year={2015}
}

@article{silva2023classifier,
  title={Classifier calibration: a survey on how to assess and improve predicted class probabilities},
  author={Silva Filho, Telmo and Song, Hao and Perello-Nieto, Miquel and Santos-Rodriguez, Raul and Kull, Meelis and Flach, Peter},
  journal={Machine Learning},
  volume={112},
  number={9},
  pages={3211--3260},
  year={2023}
}

@inproceedings{zadrozny2002transforming,
  title={Transforming classifier scores into accurate multiclass probability estimates},
  author={Zadrozny, Bianca and Elkan, Charles},
  booktitle={Proceedings of the 8th ACM SIGKDD international conference on Knowledge discovery and data mining},
  pages={694--699},
  year={2002}
}

@article{widmann2019calibration,
  title={Calibration tests in multi-class classification: A unifying framework},
  author={Widmann, David and Lindsten, Fredrik and Zachariah, Dave},
  journal={Advances in neural information processing systems},
  volume={32},
  year={2019}
}

@article{wilcox1973indices,
  title={Indices of qualitative variation and political measurement},
  author={Wilcox, Allen R},
  journal={Western Political Quarterly},
  volume={26},
  number={2},
  pages={325--343},
  year={1973}
}

@incollection{deuschel2024role,
  title={The role of uncertainty quantification for trustworthy {AI}},
  author={Deuschel, J. and Foltyn, A. and Roscher, K. and Scheele, S.},
  booktitle={Unlocking Artificial Intelligence: From Theory to Applications},
  pages={95--115},
  year={2024},
  publisher={Springer}
}

@inproceedings{guo2017calibration,
  title={On calibration of modern neural networks},
  author={Guo, Chuan and Pleiss, Geoff and Sun, Yu and Weinberger, Kilian Q},
  booktitle={International conference on machine learning},
  pages={1321--1330},
  year={2017},
  organization={PMLR}
}

\end{document}